# Machine Translation at Booking.com:
# Journey and Lessons Learned

**Pavel Levin**
Booking.com
Amsterdam
pavel.levin
@booking.com

**Nishikant Dhanuka**
Booking.com
Amsterdam
nishikant.dhanuka
@booking.com

**Maxim Khalilov**
Booking.com
Amsterdam
maxim.khalilov
@booking.com

## Abstract

We describe our recently developed neural machine translation (NMT) system and benchmark it against our own statistical machine translation (SMT) system as well as two other general purpose online engines (statistical and neural). We present automatic and human evaluation results of the translation output provided by each system. We also analyze the effect of sentence length on the quality of output for SMT and NMT systems.

## 1 Introduction

Booking.com is one of the biggest ecommerce companies in the world, offering content and serving customers in over 40 different languages. Because the need for translated content is growing very fast (in line with the overall Booking.com growth), machine translation is becoming a very attractive solution to complement the traditional human translation services.

One of the main use cases for translation at Booking.com is translating property descriptions (hotels, apartments, B&Bs, hostels, etc.) from English to any of the other supported languages. Integrating a machine translation solution would potentially dramatically increase the translation efficiency by increasing its speed and reducing the time it takes for a translated property description to appear online, as well as significantly cutting associated translation costs.

This work describes our production NMT system as well as an earlier version SMT system for two important language pairs: English-German and English-French. We benchmark the two in-house systems against each other and against two general purpose online engines (statistical and neural). Further we look at how the performance of our NMT and SMT systems varies with the sentence length.

## 2 Related work

Despite being relatively young, neural machine translation (NMT) has been quickly gaining popularity over statistical machine translation (SMT) both in academic circles and in the industry (Jean et al., 2015; Crego et al., 2016). The main reasons for this are much simpler and more elegant training pipelines, ability to address–at least in theory– some of SMT's fundamental limitations (Cho et al,. 2014; Sutskever et al., 2014) and of course as of recently the quality performance (Bojar et al., 2016; Cettolo et al., 2016; Junczys-Dowmunt et al., 2016; Wu et al., 2016).

Although there is a lot of active development in NMT research (Neubig, 2017; Sennrich et al., 2016), there have not been many demonstrations of NMT usefulness in real world scenarios (some of the exceptions include Wu et al., 2016 and Crego et al., 2016)).

In addition, in Section 4.4, we offer some empirical data related to the ongoing discussion (Cho et al., 2014; Bahdanau et al., 2015) around the NMT performance as a function of sentence length.

## 3 Experimental settings

In this section, we describe configuration and design of statistical and neural MT engines in exploration and production, as well as the data our experiments were based on.

### 3.1 Data

Experiments were conducted using internal parallel data extracted from Booking.com translation memories that contain original property descriptions in English and their translations into German and French. Note that because translation coverage vary for German and French markets, amount of training data available for English-German and English-French differ.

Basic statistics of the tokenized training corpus can be found in Table 1. Note that ASL stands for average sentence length, M stands for million, K stands for thousand.

| Language | Sent. | Words | Voc. | ASL |
|---|---|---|---|---|
| English-German | | | | |
| German | 10.5M | 171M | 845K | 16.3 |
| English | | 174M | 583K | 16.5 |
| English-French | | | | |
| French | 11.3M | 193M | 588K | 17.7 |
| English | | 188M | 581K | 16.7 |

Table 1: Statistics of the training corpora.

The development corpus was 10K segments long for NMT training and contained 5K segments for SMT tuning.

### 3.2 SMT

The SMT system we used was based on the open-source *MOSES* toolkit (Koehn et al., 2007). We followed the guidelines, as detailed on the MOSES web page[1]. Word alignment was estimated with GIZA++ tool (Och, 2003). A 5-gram target language model was estimated using the IRST LM toolkit (Federico et al., 2008). The reordering method used in the Moses-based MT systems is MSD (Tillman, 2004), coupled with a distance-based reordering.

### 3.3 NMT

Our neural machine translation system is based on *OpenNMT* (Klein et al., 2017) implementation of the global attention Sequence-to-sequence (*seq2seq*) model on words level (Luong et al., 2015). In the last few years the family of seq2seq models has been gaining significant momentum in the machine translation world. The idea behind this class of models is to encode the source sequence—usually as a fixed length vector—using some type of encoder and then to output the target sequence with a decoder, conditional on the encoded representation of the source. When trained jointly, the encoder and the decoder learn how to translate source to target (Sutskever et al., 2014).

In our system both the encoder and the decoder are long short-term memory (LSTM) recurrent neural nets (Hochreiter et al., 2017) with multiple hidden layers. The encoder LSTM reads each input sequence one token[2] at a time updating the internal representation of the sequence read so far. Those representations are essentially the LSTM hidden states. The final LSTM hidden state (after seeing the end of sequence `</s>` token in the source) is then used to initialize the decoder LSTM whose task is to generate the output sentence, again one token at a time.

In addition to the simple recurrent neural net decoder we also used an attention mechanism because letting the decoder attend to relevant parts of the source input has been shown to dramatically improve translation quality (Bahdanau et al., 2014, Luong et al., 2015). The way attention works is as follows. At each time step of generating the output we assign a probability measure over the input tokens ("alignment weights"), which we use to take the weighted average of the input hidden states and feed the resulting "context" vector as an additional input to the decoder for the current time step. The alignment weights are computed by a shallow neural network which takes the current target LSTM hidden state and each source LSTM hidden states as inputs (Luong et al., 2015).

#### 3.3.1 NMT Training

As is common (e.g. Sutskever et al., 2014; Luong et al., 2015) we use 4-layered LSTMs for both the encoder and the decoder with the vocabularies of 50K most common words for both languages (following Luong et al., 2015)[3]. All out-of-vocabulary words were encoded with a special `<unk>` symbol (following Sutskever et al., 2014 and Luong et al., 2015). Both the dimensionality of word embeddings and the LSTM hidden layer are of size 1000. Dropout (Srivastava et al., 2014) between the LSTM layers was set to 0.3. We ex-cluded any sentences of length > 50 words. The total number of parameters the model has is just over 220 million which lets us train in batches of size 250 on a single NVIDIA Tesla K80 GPU. For a typical training corpus of size 10-11M sentence pairs, each epoch takes approximately 2 days.

---

[1] http://www.statmt.org/moses/

[2] A token can be either a vocabulary word, a punctuation mark, beginning of the sentence `<s>`, end of sentence `</s>`, blank `<blank>` or out-of-vocabulary word `<unk>`.

[3] In our earlier experiments we tried using less than four layers but, as expected, got significantly worse results.

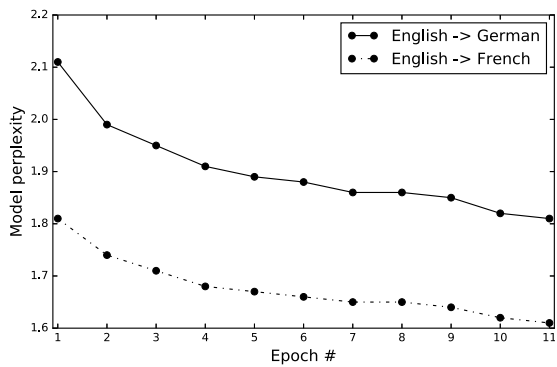

Figure 1: Model perplexity (measured on the validation set) as a function of training epoch.

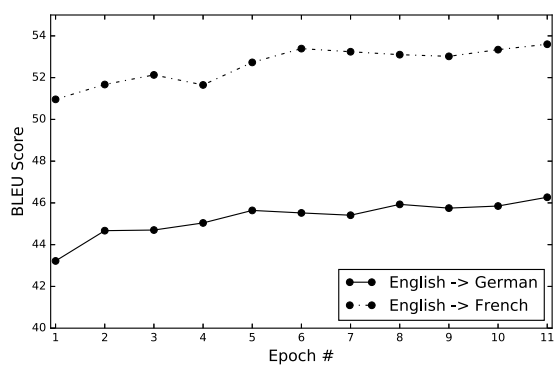

Figure 2: BLEU score (measured on the test set) as a function of training epoch.

The model parameters were fitted using normal stochastic gradient descent procedure, starting with learning rate 1, and halving it whenever the decoder perplexity on the validation set (see Figure 1) for the current epoch is not decreased. We also did BLEU score calculations on the validation set after each epoch. Our decision about when to stop was done on a case by case basis and was guided mainly by BLEU score improvements over previous epochs, manual analysis of a few hand-picked "sensitive" sentences[4] and of course our product development time constraints. Depending on a particular language pair and corpus size we would usually stop after anywhere between 5 and 13 epochs.

Because of the closed vocabulary nature of our NMT system, the output translation may contain <unk> tokens for predicted out-of-vocabulary words. To get the final version of the translation, therefore, we follow a postprocessing step in which we look at the attention score distribution of the output <unk> token over the source words and copy the one with the maximal value. Because in our use case (hotel descriptions) those words are most commonly names of places, this heuristic of copying the most probable word from the source usually works quite well in practice. Here is an example of a translated sentence with multiple out-of-vocabulary words:

| | |
|---|---|
| Source | *Offering a restaurant, Hodor Eco-lodge is located in Winterfell.* |
| Human Translation | *Das Hodor Eco-Lodge begrüßt Sie in Winterfell mit einem Restaurant.* |
| Raw Output | *Das <unk><unk> in <unk> bietet ein Restaurant.* |
| Output with <unk> replaced | *Das Hodor Eco-lodge in Winterfell bietet ein Restaurant.* |

## 4 Evaluation

We compared translation quality delivered by 4 MT systems: in-house *SMT* and *NMT* as described in the previous section, as well as statistical and neural online general purpose engines (*SGPMT* and *NGPMT*) trained on the general domain data.

### 4.1 Automatic evaluation

We used BLEU metric as the primary automatic metric of translation quality evaluation. BLEU (Papineni et al., 2002) shows the number of words shared between MT output and human-made reference, benefiting sequential words and penalizing very short translations. BLEU scores were calculated on the basis of truecased and detokenized test datasets of 10K segments and one reference translation. The evaluation conditions were case-sensitive and included punctuation marks.

In our analysis of the effect of the sentence length on machine translation quality (Section 4.4) we also use Word Error Rate (WER). WER is a variation of the word-level Levenshtein distance measuring the distance between the target and the reference sentences by counting the insertions, deletions and substitutions necessary to go from one to the other.

---

[4] For example in some languages "The neighbourhood is very nice and safe" is often translated to mean "There is a safe installed in this very nice neighbourhood" during the early learning stage because the word *safe* is very often used to mean *a safe box* in our property descriptions.

### 4.2 Manual evaluation

We validated results of our findings with human *Adequacy-Fluency (AF)* evaluation applying a 4-level scale to both Adequacy and Fluency as described in the TAUS Adequacy/Fluency Guidelines[5].

Evaluators (3 per language), which are native speakers of the target language, were provided with the original text in English and the MT hypotheses. They were asked to assess the quality of 150 randomly selected lines from the test corpus translated by the four MT systems under consideration. The evaluators were not aware of which system produced which hypothesis.

### 4.3 Evaluation results

Table 2 presents BLEU and AF scores of our benchmarking experiment. Figures 3 and 4 shows the human evaluation results.

| Translation | BLEU | Adequacy | Fluency |
|---|---|---|---|
| English-German | | | |
| SMT | 35.24 | 3.62 | 3.15 |
| NMT | 45.64 | 3.90 | 3.78 |
| SGPMT | 27.63 | 3.57 | 3.37 |
| NGPMT | 31.45 | 3.65 | 3.57 |
| Human | – | 3.96 | 3.82 |
| English-French | | | |
| SMT | 35.80 | 3.40 | 3.28 |
| NMT | 52.73 | 3.67 | 3.40 |
| SGPMT | 30.25 | 3.32 | 3.31 |
| NGPMT | 32.18 | 3.78 | 3.41 |
| Human | – | 3.70 | 3.75 |

Table 2: Evaluation results.

We observed that:

- According to the BLEU scores, NMT consistently outperform all other engines with a significant margin;

- Both neural systems (NMT and NGPMT) consistently outperform their statistical counterparts (SMT and SGPMT) according to both automatic and manual metrics;

- The performance of general purpose engines is worse than that of the in-house engines in case of English-German in terms of both BLEU and A/F scores, while in case of English-French, there is a mismatch between BLEU and adequacy score. In the latter case, NGPMT outperformed all other engines and surprisingly human translators in terms of adequacy, which may be an artifact of the small sample size, as well as the subjectivity of the metric itself.

- The fluency performance of the NMT engines is not far from human level for English-German, while for English-French adequacy delivered by both neural engines (in-house NMT and NGPMT) is approximately at the human translation level.

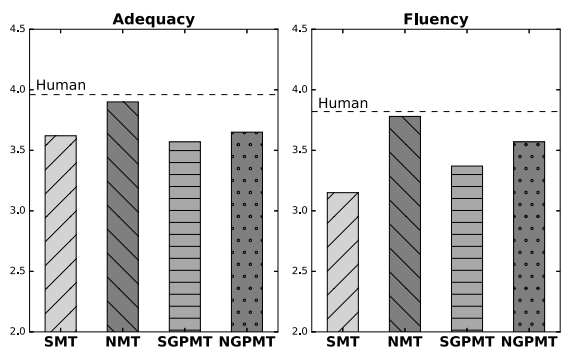

Figure 3: AF results for English-German for the four systems and a human translation benchmark.

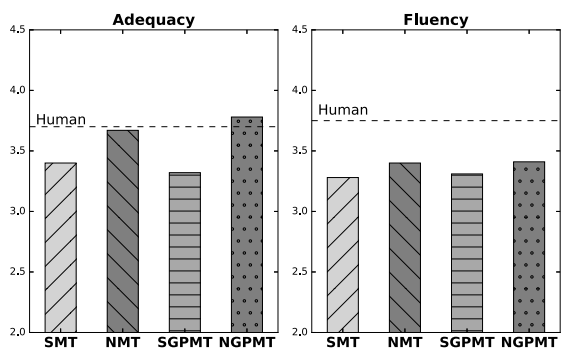

Figure 4: AF results for English-French for the four systems and a human translation benchmark.

---
[5]https://www.taus.net/academy/
best-practices/evaluate-best-practices/
adequacy-fluency-guidelines

## 4.4 Sentence length analysis

Multiple studies (Cho et al., 2014a) find that translation quality drops significantly when NMT translates long sentences. The primary cause being that, for longer sentences, the fixed-size vector representations of source sentences by encoder struggles to capture all cues for decoder to generate appropriate translations. Attention mechanism helps to combat this problem to a certain extent by selectively focusing on relevant parts of the source sentence while translating, instead of just relying on a fixed vector representation. There are other approaches as well, for example breaking long sentence into shorter phrases before translation (Pouget-Abadie et al., 2014). We were interested to see the correlation between sentence length and the machine translation quality in our data, particularly whether SMT outperforms NMT for longer sentences.

We segmented our tokenized test corpus into 10 bins according to lengths of the source sentences. Each bin contained roughly 1,000 sentences. We then ran BLEU score and negative word error rate evaluation separately on each of the 10 batches. Results are displayed in Figures 5-8.

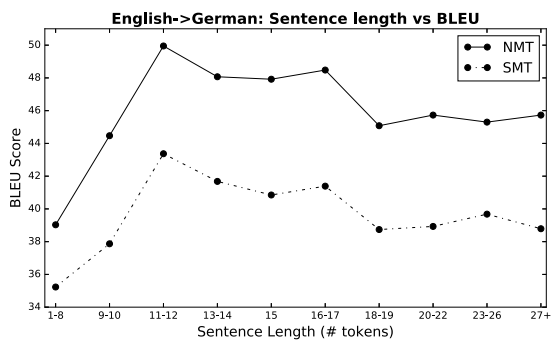

Figure 5: Sentence Length vs. Quality (BLEU) for SMT & NMT in English-German translation.

Our observation is two-fold:

- both systems, roughly followed the same trend. Quality was low for very small sentences i.e. 1-8 tokens, then increased with the length as the context helped in translation, but reached a peak soon around 11-17 tokens, and thereafter degraded for longer sentences;

- even for longer sentences though performance degraded, our NMT system outperformed SMT.

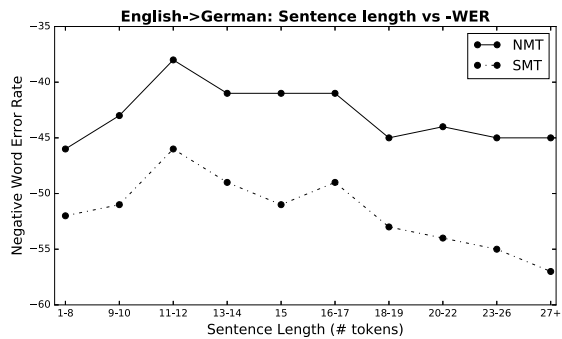

Figure 7: Sentence Length vs. Quality (-WER) for SMT & NMT in English-German translation.

We ran the same experiment on English to French translations, and observed very similar trends (See Figure 6).

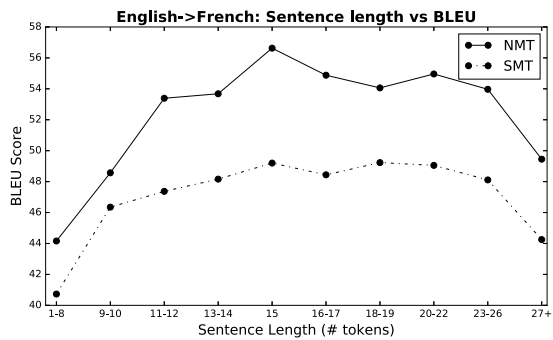

Figure 6: Sentence Length vs. Quality (BLEU) for SMT & NMT in English-French translation.

We used WER as a secondary metric to validate the results of BLEU analysis which could be biased for shorter sentences. We report negative WER to make this into a precision measure.

As can be seen in Figures 7 and 8, the results are very similar to those in Figures 5 and 6. This further corroborates our observations outside the constraints of the BLEU score.

## 5 Conclusions and future work

The main three findings of this study are: (1) neural MT technology consistently outperforms statistical; (2) in case of German, in-house NMT is also better than online general purpose engines in our application; (3) fluency of NMT is close to human translation level; and (4) in our application the relative performance of NMT against SMT does not degrade with increased sentence length.

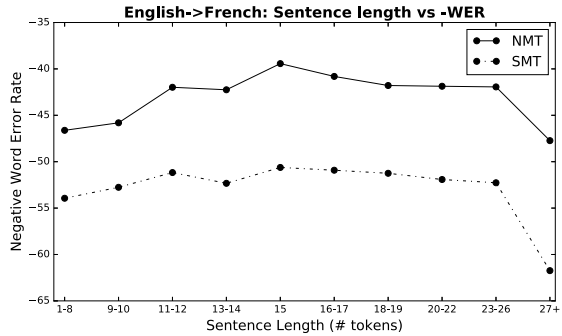

Figure 8: Sentence Length vs. Quality (-WER) for SMT & NMT in English-French translation.

Our future research directions include further improving our in-house NMT system in two important ways. The first one is the improved treatment of unknown and rare words which are particularly important to us because of a large number of named entities in our corpora, such as landmark or hotel names. The problem becomes even bigger with user generated content which may contain many misspellings and abbreviations. The second direction of research is improving our ability to identify business sensitive translation errors (e.g. "free" being translated to "available").